\documentclass[12pt]{elsarticle}

\usepackage[letterpaper,top=1.2in,bottom=1.5in,left=1.0in,right=1.0in]{geometry}%

\usepackage{amssymb,mathrsfs,amsmath}
\usepackage{MnSymbol,bbding,pifont}
\usepackage{extarrows}
\usepackage{soul}
\usepackage{hyperref}
\hypersetup{hidelinks,
	colorlinks=true,
	allcolors=black,
	pdfstartview=Fit,
	breaklinks=true}

\usepackage{color}
\usepackage{courier}
\usepackage{booktabs}
\usepackage{makecell}
\usepackage{multirow}
\usepackage{siunitx}
\usepackage{bm}
\usepackage[figuresright]{rotating}
\usepackage[dvipsnames]{xcolor}
\usepackage[most]{tcolorbox}
\usepackage[toc,page,header]{appendix}
\usepackage{minitoc}

\usepackage[switch]{lineno}
\renewcommand\appendix{\par
    \setcounter{section}{0}
    \setcounter{subsection}{0}
    \gdef\thesection{\Alph{section}}}

{}
{}
\newtheorem{remark}{\bf Remark}{}
{}

\usepackage{standalone}
\usepackage{tikz}
\usetikzlibrary{patterns}
\usepackage{graphicx}
\usepackage{subcaption}

\usepackage{algorithm}
\usepackage{algorithmic}

\journal{XXXXXXXXXXX}

\begin{document}
\doparttoc % Tell to minitoc to generate a toc for the parts
\faketableofcontents % Run a fake tableofcontents command for the partocs
\parttoc % Insert the document TOC
% \begin{linenumbers}
\begin{frontmatter}
 


% \title{Proportion-Integral-Derivative Accelerated Optimizer for Deep Learning}

\title{Distributed physics-informed neural networks via domain decomposition for fast flow reconstruction}

% \author[inst1]{Name\corref{cor1}}
% % \ead[url]{home page}
% \cortext[cor1]{Corresponding author}

\author[A]{Yixiao Qian}
\author[B]{Jiaxu Liu\corref{cor1}}\ead{jiaxuliu@zju.edu.cn}
\author[A]{Zewei Xia}
\author[C]{Song Chen}
\author[A,D]{Chao Xu}
\author[A,D]{Shengze Cai\corref{cor1}}\ead{shengze\_cai@zju.edu.cn}

\cortext[cor1]{Corresponding author}

\affiliation[A]{organization={College of Control Science and Engineering},%Department and Organization
            addressline={Zhejiang University}, 
            city={Hangzhou},
            country={China},
            }
\affiliation[B]{organization={School of Mathematical Sciences},%Department and Organization
            addressline={Zhejiang University}, 
            city={Hangzhou},
            country={China},
            }
\affiliation[C]{organization={Department of Mathematics},
            addressline={National University of Singapore}, 
            country={Singapore},
            }
\affiliation[D]{organization={State Key Laboratory of Industrial Control Technology},
            addressline={Zhejiang University}, 
            city={Hangzhou},
            country={China},
            }

\begin{abstract}
Physics-Informed Neural Networks (PINNs) offer a powerful paradigm for flow reconstruction, seamlessly integrating sparse velocity measurements with the governing Navier-Stokes equations to recover complete velocity and latent pressure fields.
However, scaling such models to large spatiotemporal domains is hindered by computational bottlenecks and optimization instabilities. 
In this work, we propose a robust distributed PINNs framework designed for efficient flow reconstruction via spatiotemporal domain decomposition.
A critical challenge in such distributed solvers is pressure indeterminacy, where independent sub-networks drift into inconsistent local pressure baselines. 
We address this issue through a reference anchor normalization strategy coupled with decoupled asymmetric weighting. 
By enforcing a unidirectional information flow from designated master ranks where the anchor point lies to neighboring ranks, our approach eliminates gauge freedom and guarantees global pressure uniqueness while preserving temporal continuity.
Furthermore, to mitigate the Python interpreter overhead associated with computing high-order physics residuals, we implement a high-performance training pipeline accelerated by CUDA graphs and JIT compilation. 
Extensive validation on complex flow benchmarks demonstrates that our method achieves near-linear strong scaling and high-fidelity reconstruction, establishing a scalable and physically rigorous pathway for flow reconstruction and understanding of complex hydrodynamics.
\end{abstract}


% \begin{highlights}
% \item Distributed PINNs with spatiotemporal domain decomposition for fast flow reconstruction from sparse velocity data.
% \item Reference anchor normalization and decoupled asymmetric weighting eliminate pressure gauge indeterminacy in distributed training.
% \item CUDA Graphs/JIT acceleration improves throughput and enables near-linear strong scaling on 2D/3D flow benchmarks.
% \end{highlights}


\end{frontmatter}


\section{Introduction}
\label{sec:intro}

The reconstruction of high-resolution fluid flow fields constitutes a fundamental task in fluid dynamics, playing a pivotal role in both scientific discovery and practical engineering applications. 
Accurately recovering continuous velocity and pressure fields from limited observations is essential for elucidating complex flow mechanisms, such as turbulence evolution \cite{zaki2021limited,fukami2019super} and the dynamics of coherent structures \cite{holmes2012turbulence,wu2017transition}. 
Beyond fundamental physics, the demand for high-fidelity flow data extends across a broad spectrum of disciplines, including chemical mixing \cite{wang20243}, environmental remote sensing \cite{shen2015missing}, and experimental fluid mechanics \cite{cai2024physics}. 
While optical diagnostic techniques like Particle Image Velocimetry (PIV) \cite{adrian2011particle} and Particle Tracking Velocimetry (PTV) \cite{dabiri2019particle} serve as the cornerstone of experimental observation, they are frequently constrained by limited spatial resolution and inevitable measurement noise.
Consequently, developing robust techniques to infer dense spatiotemporal information from these sparse measurements has attracted significant attention, aiming to bridge the critical gap between discrete experimental observations and the comprehensive field descriptions required for in-depth analysis.

Conventionally, the reconstruction of continuous flow fields from discrete measurements has relied on numerical interpolation \cite{hildebrand1987introduction} or traditional data assimilation (DA) techniques \cite{wang2021state}. 
Interpolation methods, such as Kriging or splines \cite{zhou20193,sirakov2002interpolation}, typically assume spatial smoothness, which often leads to significant errors when resolving high-frequency turbulent fluctuations. 
To enforce physical consistency, DA approaches—including adjoint-based optimization \cite{mons2019kriging,buchta2021observation}, nudging schemes \cite{clark2018inferring,clark2020synchronization}, and four-dimensional variational (4DVar) methods \cite{he2024four,li2023unsteady}—integrate observational data with numerical simulations. 
Although these methods can theoretically recover accurate flow states, they are computationally prohibitive and entail complex implementation procedures, particularly for high-Reynolds-number flows \cite{wang2023unified}. 
In recent years, deep learning-based techniques, such as Convolutional Neural Networks (CNNs) \cite{fukami2019super,cai2019particle}, Generative Adversarial Networks (GANs) \cite{xie2018tempogan,buzzicotti2021reconstruction}, and Diffusion Models \cite{shan2024pird, shu2023physics}, have shown promise in efficiently mapping low-resolution inputs to high-resolution outputs. 
However, these ``black-box'' models generally necessitate massive amounts of high-fidelity labeled training data—which are rarely available in experiments—and may generate non-physical artifacts due to the lack of explicit governing constraints \cite{wang2022dense}.

To address these challenges, Physics-Informed Neural Networks (PINNs) have emerged as a transformative paradigm that seamlessly integrates governing physical laws with observational data \cite{raissi2019physics, cai2021physics, chu2024flow}. 
By embedding the Navier-Stokes equations directly into the loss function, PINNs regularize the training process, enabling the reconstruction of flow fields from sparse and noisy measurements without requiring dense ground truth data. 
PINNs have been extensively adapted to solve incompressible flows across flow regimes, ranging from laminar benchmarks to turbulent channel flows \cite{rao2020physics,jin2021nsfnets}. 
Subsequent studies have further augmented this framework to address increasingly complex physical phenomena: Zhang et al. \cite{zhang2020frequency} utilized Fourier feature mappings to capture high-frequency dynamics; Cheng et al. \cite{cheng2021deep} demonstrated efficacy in fluid-structure interactions; and Ouyang et al. \cite{ouyang2023reconstruction} extended the methodology to multiphase flows. 
Crucially, the practical utility of PINNs in experimental fluid mechanics has been rigorously validated. 
Raissi et al. \cite{raissi2020hidden} pioneered the concept of ``hidden fluid mechanics'' to infer velocity and pressure solely from flow visualizations. 
Building on this, recent works have successfully applied PINNs to tomographic PTV data \cite{cai2024physics} and noisy experimental measurements \cite{clark2023reconstructing}, recovering high-fidelity instantaneous pressure fields that are otherwise challenging to measure directly.

However, despite these advancements, the standard PINNs architecture faces significant limitations when scaling to large spatiotemporal domains. 
Relying on a single monolithic network to approximate the entire flow field inevitably leads to the ``spectral bias'' phenomenon, where the network struggles to capture high-frequency turbulent fluctuations while simultaneously maintaining global low-frequency consistency. 
As domain size and flow complexity increase, the centralized optimization strategy incurs prohibitive computational costs and memory bottlenecks, often leading to optimization pathologies and poor convergence \cite{wang2022and}. 

To mitigate these computational bottlenecks and spectral bias, Domain Decomposition (DD) strategies have been integrated into the PINNs framework, enabling parallelized training across distributed computing resources. 
Pioneering works have established the theoretical foundations for this approach: Jagtap and Karniadakis \cite{jagtap2020extended} introduced Extended PINNs (XPINNs), a generalized space-time decomposition framework applicable to nonlinear PDEs. 
Complementary to this, conservative PINNs (cPINNs) were proposed to handle conservation laws by enforcing flux continuity across subdomains \cite{jagtap2020conservative}.
Building upon these concepts, Shukla et al. \cite{shukla2021parallel} developed a distributed MPI-based framework for both cPINNs and XPINNs, demonstrating significant computational speedup and the ability to optimize hyperparameters independently within each subdomain.
Similarly, Meng et al. \cite{meng2020ppinn} proposed Parareal PINNs (PPINN) to parallelize temporal evolution for time-dependent problems, while Ye et al. \cite{ye2025spatial} demonstrated the efficacy of overlapping DD for solving various differential equations.
However, despite their algorithmic contributions, these existing approaches have predominantly focused on two-dimensional benchmarks or well-posed \textit{forward} problems where boundary conditions are fully prescribed. 
A dedicated framework for efficient and robust \textit{flow reconstruction}—an ill-posed inverse problem inferred strictly from sparse experimental observations—remains largely unexplored. 
Crucially, standard distributed methods lack specific mechanisms to handle the pressure indeterminacy inherent in experimental data assimilation (where pressure boundaries are unknown), rendering them susceptible to numerical instability when scaled to complex three-dimensional flow fields.

In this work, we present a robust distributed physics-informed neural network framework designed specifically for large-scale flow field reconstruction. 
By addressing the unique challenges of solving inverse problems without pressure boundary conditions, our approach bridges the gap between scalable computing and high-fidelity experimental data assimilation. 
The main contributions of this study are summarized as follows:

\begin{itemize}
    \item \textbf{Distributed PINNs framework for fast flow reconstruction:}
    We propose a distributed PINN framework based on spatiotemporal domain decomposition, enabling parallel training of local experts for large-scale flow reconstruction from sparse velocity measurements.

    \item \textbf{Resolution of pressure indeterminacy:}
    We address pressure indeterminacy in distributed Navier-Stokes reconstruction via \textit{Reference Anchor Normalization} and \textit{Decoupled Asymmetric Weighting}, preventing inter-domain pressure-offset drift without requiring pressure boundary conditions.

    \item \textbf{High-throughput implementation:}
    We integrate CUDA Graphs and Just-In-Time (JIT) compilation \cite{bafghi2023pinnstorch} to reduce Python overhead in high-order derivative evaluation and improve per-GPU throughput.

    \item \textbf{Strong-scaling validation:}
    We demonstrate near-linear strong scaling and accurate reconstructions on canonical 2D and 3D benchmarks, including the lid-driven cavity and cylinder wakes.
\end{itemize}

The remainder of this paper is organized as follows. 
Section \ref{sec:method} outlines the methodological framework, beginning with a formulation of PINNs for flow reconstruction. 
We then detail the proposed algorithm based on domain decomposition, specifically elaborating on the anchor-based pressure alignment mechanism, the CUDA Graph-accelerated optimization pipeline, and heuristic strategies for loss weighting. 
In Section \ref{sec:num_results}, we validate the accuracy and scalability of our approach using high-fidelity CFD data across canonical benchmarks, including the 2D lid-driven cavity, 2D flow past a cylinder, and the 3D cylinder wake. Finally, Section \ref{sec:conclusion} concludes this
paper.

\section{Methodology}
\label{sec:method}

\subsection{Physics-informed neural networks for flow field reconstruction}

Flow field reconstruction aims to recover continuous, high-fidelity velocity fields $\mathbf{u}(\mathbf{x}, t)$ and pressure field $p(\mathbf{x}, t)$ within the spatiotemporal domain $\Omega \times T$ from sparse experimental observations acquired via PIV/PTV. 
In this framework, PINNs serve as a mesh-free post-processing technique that assimilates experimental data while rigorously enforcing physical conservation laws.
Let the sparse velocity measurements obtained from PIV/PTV be denoted as:
\begin{equation}
    \mathcal{D}_{\text{obs}}: \{\mathbf{x}^n_{\text{obs}}, t^n_{\text{obs}}, \mathbf{u}^n_{\text{obs}}\}_{n=1}^{N_{\text{obs}}},
\end{equation}
where $\mathbf{x} \in \Omega \subset \mathbb{R}^d$ and $t \in T$ represent the spatial and temporal coordinates, respectively; $\mathbf{u} \in \mathbb{R}^d$ denotes the velocity vector; and $N_{\text{obs}}$ is the total number of observed data points in $\mathcal{D}_{\text{obs}}$.

We approximate the flow field variables using a neural network parametrized by $\boldsymbol{\theta}$:
\begin{equation}
    (\mathbf{u}_{\mathcal{NN}}, p_{\mathcal{NN}}) = \mathcal{NN}(\mathbf{x}, t; \boldsymbol{\theta}), \quad \mathbf{x} \in \Omega, \quad t \in T.
\end{equation}
To fit the experimental observations, the network is trained via supervised learning by minimizing the mean squared error (MSE) between the predicted and observed velocities:
\begin{equation}
    \mathcal{L}_{\text{obs}}(\boldsymbol{\theta}) := \frac{1}{N_{\text{obs}}} \sum_{n=1}^{N_{\text{obs}}} \| \mathbf{u}_{\text{obs}}^n - \mathbf{u}_{\mathcal{NN}}(\mathbf{x}_{\text{obs}}^n, t_{\text{obs}}^n; \boldsymbol{\theta})\|_2^2.
\end{equation}

Given the sparsity of observations, reconstruction relying solely on $\mathcal{D}_{\text{obs}}$ may yield poor generalization in regions lacking measurement data. To address this, physical constraints are enforced at a set of collocation points, denoted as $\mathcal{D}_{\text{PDE}}$. These points are randomly sampled from the computational domain $\Omega \times T$:
\begin{equation}
    \mathcal{D}_{\text{PDE}} := \{\mathbf{x}_{\text{PDE}}^n, t_{\text{PDE}}^n\}_{n=1}^{N_{\text{PDE}}}.
\end{equation}
The corresponding physics-informed loss is defined as the mean squared residuals of the governing equations evaluated at these sampled points:
\begin{equation}
    \mathcal{L}_{\text{PDE}}(\boldsymbol{\theta}) := \frac{1}{N_{\text{PDE}}}\sum_{n=1}^{N_{\text{PDE}}} \|\mathbf{r}(\mathbf{x}_{\text{PDE}}^n, t^n_{\text{PDE}}; \boldsymbol{\theta})\|_2^2,
\end{equation}
where $\mathbf{r}$ represents the residual vector.
In this work, we consider the incompressible Navier-Stokes equations
\begin{equation}
      \frac{\partial \mathbf{u}}{\partial t} + (\mathbf{u} \cdot \nabla)\mathbf{u} = -\nabla p + \frac{1}{\mathrm{Re}} \nabla^2 \mathbf{u},
      \quad \nabla \cdot \mathbf{u} = 0.
\end{equation}
For three-dimensional flows where $\mathbf{u} = [u, v, w]^\top$,
the explicit component-wise residuals $\mathbf{r} = [r_1, r_2, r_3, r_4]^\top$ are defined as:
\begin{align}
      r_1 := &u_t + (u u_x + v u_y + w u_z) + p_x - \tfrac{1}{\mathrm{Re}}(u_{xx}+u_{yy}+u_{zz}),\\
      r_2 := &v_t + (u v_x + v v_y + w v_z) + p_y - \tfrac{1}{\mathrm{Re}}(v_{xx}+v_{yy}+v_{zz}),\\
      r_3 := &w_t + (u w_x + v w_y + w w_z) + p_z - \tfrac{1}{\mathrm{Re}}(w_{xx}+w_{yy}+w_{zz}),\\
      r_4 := &u_x + v_y + w_z.
\end{align}
Minimizing these residuals (i.e., driving $\|\mathbf{r}\| \to 0$) constrains the neural network predictions to lie on the manifold of physically valid solutions, thereby effectively regularizing the reconstruction in spatiotemporally sparse regions.

The optimal network parameters $\boldsymbol{\theta}^*$ are determined by minimizing a composite loss function $\mathcal{L}$, defined as a weighted sum of the observation and PDE losses:
\begin{equation}
    \mathcal{L}(\boldsymbol{\theta}) := \lambda_{\text{obs}} \mathcal{L}_{\text{obs}} + \lambda_{\text{PDE}} \mathcal{L}_{\text{PDE}},
\end{equation}
where $\lambda_{\text{obs}}$ and $\lambda_{\text{PDE}}$ are weighting coefficients that balance data fidelity with physical constraints. By optimizing this objective, the network simultaneously interpolates the velocity field and infers the pressure distribution, ensuring the reconstructed flow is both consistent with measurements and physically compliant.

\subsection{Physics-informed neural networks with domain decomposition}

Reconstructing large-scale spatiotemporal fields with a single PINN can be computationally expensive and often yields suboptimal accuracy. 
A monolithic network lacks the capacity to capture high-frequency details uniformly, often leading to oversmoothed results in turbulent regions. 
Additionally, the sheer volume of training data required for large domains results in excessively long training durations.
To resolve these issues, we implement a domain decomposition (DD) strategy. 
By decomposing the global problem into smaller, manageable sub-problems, we leverage parallel computing to achieve scalable and fast reconstruction. 

Specifically, we partition the global domain $\Omega \times T$ into a grid of $K \times M$ disjoint sub-domains $\{\Omega_k^{\mathrm{int}} \times T_m^{\mathrm{int}}\}$, where $k \in \{1, \dots, K\}$ denotes the spatial index and $m \in \{1, \dots, M\}$ denotes the temporal index.
Instead of utilizing a single global network, we assign a separate, independent neural network $\mathcal{NN}_{k,m}(\mathbf{x}, t; \boldsymbol{\theta}_{k,m})$—referred to as a ``local expert''—to each spatiotemporal subdomain.
This approach allows each local expert to focus on learning localized flow features, significantly reducing optimization complexity. Since the local networks are computationally independent, they can be trained simultaneously on distributed computing resources.

\begin{figure}[htbp]
    \centering
    \includegraphics[width=0.75\textwidth]{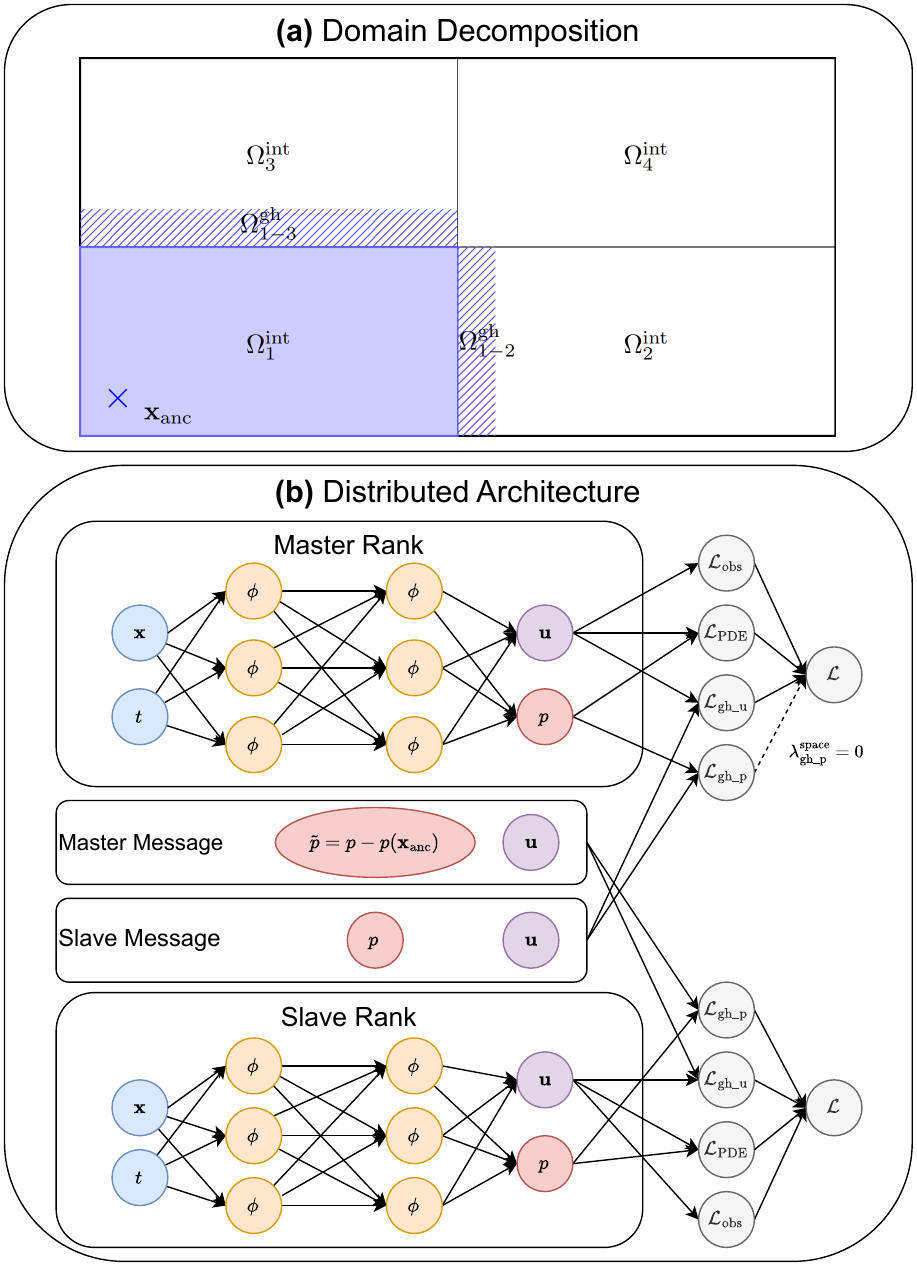}
    \caption{Illustration of the domain decomposition and distributed training architecture. 
    \textbf{(a)} Spatial domain decomposition in a fixed time interval: the global domain $\Omega$ is partitioned into four sub-domains, $\Omega_1^{\text{int}}$ through $\Omega_4^{\text{int}}$. The ghost layer $\Omega_1^{\text{gh}} = \Omega_{1-2}^{\text{gh}} \cup \Omega_{1-3}^{\text{gh}}$ is introduced by extending $\Omega_1^{\text{int}}$ to enforce continuity across the interfaces. 
    The anchor point $\mathbf{x}_{\text{anc}}$ identifies the master rank (here $\Omega_1$), and other sub-domains align their pressure gauge to this reference. 
    \textbf{(b)} Distributed PINN training: both master and slave ranks learn $\mathcal{NN}:(\mathbf{x},t)\mapsto(\mathbf{u},p)$, and the interior losses $\mathcal{L}_{\text{obs}}$ and $\mathcal{L}_{\text{PDE}}$ are computed using the raw network outputs (thus independent of the pressure gauge). 
    Anchor normalization is applied only when the master transmits ghost-layer pressure to its neighbors, using $\tilde{p}=p(\mathbf{x},t)-p(\mathbf{x}_{\text{anc}},t)$. 
    For stability, the master ignores the \emph{spatial} ghost pressure loss within the same time interval (i.e., $\lambda_{\text{gh\_p}}^{\text{space}}=0$ for master ranks), while retaining the \emph{temporal} ghost pressure loss to preserve continuity across consecutive time intervals.}
    \label{fig:domain_decomposition}
\end{figure}

However, training isolated sub-networks may lead to physical discontinuities across boundaries. To ensure global solution continuity, we introduce \textit{ghost layers} extending into neighboring sub-domains. 
For each spatiotemporal subdomain $\Omega_{k,m}^{\text{int}} := \Omega_k^{\text{int}} \times T_m^{\text{int}}$, we extend its boundaries in both space and time to form an extended computational domain $\Omega_{k,m} = \Omega^{\text{int}}_{k,m} \cup \Omega^{\text{gh}}_{k,m}$.
The ghost layer $\Omega_{k,m}^{\text{gh}}$ is defined as the intersection between this extended region and the interiors of neighboring subdomains.
As a result, $\Omega_{k,m}^{\text{gh}}$ generally consists of multiple components (one per spatial and/or temporal neighbor).
Figure \ref{fig:domain_decomposition}(a) illustrates the spatial partitioning within a fixed time interval, explicitly highlighting the resultant spatial interfaces between adjacent sub-domains.
We distinguish between two types of interfaces:
\begin{itemize}
    \item \textbf{Spatial Interfaces}: Ghost layers between adjacent spatial domains $\Omega_{k,m}^{\text{int}}$ and $\Omega_{j,m}^{\text{int}}$ within the same time interval $T_m$.
    \item \textbf{Temporal Interfaces}: Ghost layers between consecutive time intervals $T_m$ and $T_{m+1}$ for the same spatial domain $\Omega_k$.
\end{itemize}
To enforce continuity across these interfaces, we construct a specific ghost layer dataset for each subdomain, denoted as $\mathcal{D}_{\text{gh}}$. 
This dataset consists of residual points uniformly sampled from the union of all valid spatial and temporal ghost layers associated with the $(k,m)$-th local expert:
\begin{equation}
    \mathcal{D}_{\text{gh}} := \left\{ (\mathbf{x}_{\text{gh}}^i, t_{\text{gh}}^i) \in \Omega_{k,m}^{\text{gh}} \right\}_{i=1}^{N_{\text{gh}}}.
\end{equation}

During training, information including $\mathbf{u}, p$ is exchanged between neighboring networks at fixed intervals. 
We then enforce consistency by minimizing the discrepancy between the local expert's predictions and the cached values from its neighbors on the ghost dataset. 
The consistency losses for velocity and pressure are defined as:
\begin{align}
    \mathcal{L}_{\mathrm{gh\_u}}^{k,m} &:= \frac{1}{N_{\text{gh}}}\sum_{(\mathbf{x}, t) \in \mathcal{D}_{\text{gh}}} \|\mathbf{u}_{\mathcal{NN}_{k,m}}(\mathbf{x}, t) - \mathbf{u}_{\text{nbr}}(\mathbf{x}, t)\|_2^2,\\
    \mathcal{L}_{\mathrm{gh\_p}}^{k,m} &:= \frac{1}{N_{\text{gh}}}\sum_{(\mathbf{x}, t) \in \mathcal{D}_{\text{gh}}} |p_{\mathcal{NN}_{k,m}}(\mathbf{x}, t) - p_{\text{nbr}}(\mathbf{x}, t)|^2,
    \label{eq:loss-ghost-pressure}
\end{align}
where $\mathbf{u}_{\text{nbr}}$ and $p_{\text{nbr}}$ denote the predictions provided by the corresponding spatial or temporal neighboring experts. By minimizing these terms, the independent networks are constrained to converge toward a globally continuous and physically consistent solution.

A fundamental challenge in distributed flow reconstruction is \textit{pressure indeterminacy}. 
Since the incompressible Navier-Stokes equations depend on pressure solely through its gradient $\nabla p$, the solution is unique only up to an arbitrary time-dependent additive function $C(t)$ (i.e., gauge freedom). 
In a monolithic network, this is typically resolved by pinning a single point or enforcing a zero-mean condition. 
However, in a distributed framework, each independent local expert infers its own local pressure baseline. 
Naively unifying these floating baselines via the standard consistency loss (\ref{eq:loss-ghost-pressure}) induces a pathological ``seesaw'' instability, where adjacent networks continuously adjust their offsets in response to one another without converging to a unified global reference.

To resolve this, we propose a \textbf{reference anchor normalization} strategy combined with \textbf{decoupled asymmetric weighting}.
We designate a specific spatial location $\mathbf{x}_{\text{anc}}$ as the global anchor. 
The subset of local experts covering this location across all time intervals is identified as \textit{master ranks}, denoted as $\mathcal{M} = \{\mathcal{NN}_{\text{anc}, m}\}_{m=1}^M$.
For these master ranks, we enforce a hard normalization constraint specifically during the evaluation of inter-domain consistency. 
When a master rank broadcasts its pressure field to ghost regions, it subtracts the instantaneous pressure value at the anchor point:
\begin{equation}
    \tilde{p}_{\mathcal{NN}_{\text{anc}, m}}(\mathbf{x}, t) = p_{\mathcal{NN}_{\text{anc}, m}}(\mathbf{x}, t) - p_{\mathcal{NN}_{\text{anc}, m}}(\mathbf{x}_{\text{anc}}, t).
\end{equation}
This operation effectively pins the broadcasted pressure field to zero at $\mathbf{x}_{\text{anc}}$, strictly eliminating the gauge function $C(t)$ in the ghost layers.
Crucially, this normalization applies \textit{only} to the ghost consistency checks ($\mathcal{L}_{\mathrm{gh\_p}}$). 
The physics residual calculation ($\mathcal{L}_{\text{PDE}}$) continues to utilize the raw network output $p_{\mathcal{NN}}$, leveraging the fact that Navier-Stokes residuals are invariant to additive constants. 
This design decouples the physical constraints from the alignment constraints, avoiding unnecessary optimization complexity.

To prevent the stabilized master ranks from being affected by the floating baselines of their neighbors, we introduce an asymmetric weighting strategy that decouples spatial and temporal interfaces:
\begin{enumerate}[(1)]
    \item \textbf{Spatial Asymmetry:} For the master ranks, the weighting coefficient for the \textit{spatial} ghost pressure loss is strictly set to zero (i.e., $\lambda_{\text{gh\_p}}^{\text{space}} = 0$). This imposes a unidirectional constraint: the master rank propagates the normalized pressure baseline to its spatial neighbors (non-master ranks) without adjusting its own parameters to minimize interfacial discrepancies.
    \item \textbf{Temporal Continuity:} Conversely, for \textit{all} ranks (including master ranks), the weighting coefficient for the \textit{temporal} ghost pressure loss remains positive (i.e., $\lambda_{\text{gh\_p}}^{\text{time}} > 0$). This preserves the temporal continuity of the solution, ensuring that the pressure field remains continuous across consecutive time intervals and mitigating potential discontinuities that could arise from the discrete enforcement of the spatial anchor.
\end{enumerate}

The final local objective function $\mathcal{L}_{k,m}$ is thus formulated as:
\begin{equation}
    \label{eq:entire-loss-func}
    \mathcal{L}_{k,m}(\boldsymbol{\theta}_{k,m}) = 
    \lambda_{\text{obs}} \mathcal{L}_{\text{obs}} 
    + \lambda_{\text{PDE}} \mathcal{L}_{\text{PDE}} 
    + \lambda_{\text{gh\_u}} \mathcal{L}_{\text{gh\_u}} 
    + \lambda_{\text{gh\_p}}^{\text{space}} \mathcal{L}_{\text{gh\_p}}^{\text{space}}
    + \lambda_{\text{gh\_p}}^{\text{time}} \mathcal{L}_{\text{gh\_p}}^{\text{time}},
\end{equation}
where $\lambda_{\text{gh\_p}}^{\text{space}}$ is set to 0 for master ranks and $>0$ otherwise, while $\lambda_{\text{gh\_p}}^{\text{time}}$ is universally active. 
No explicit anchor loss term is needed in (\ref{eq:entire-loss-func}), as the constraint is strictly enforced via the definition of $\tilde{p}$ in the ghost terms.
This hierarchical formulation ensures that the pressure field is globally unique, spatially synchronized, and temporally continuous.
Algorithm \ref{alg:dpinn} outlines the complete training workflow, including domain decomposition, information exchange, and the parallel optimization of local experts. 
It explicitly distinguishes the roles of master and slave ranks: master ranks broadcast normalized pressure messages and set $\lambda_{\text{gh\_p}}^{\text{space}}$ to zero to act as fixed references. 
Meanwhile, slave ranks employ the normalized data to correct their pressure baselines. 
This process ensures global consistency while each network minimizes the loss in Equation (\ref{eq:entire-loss-func}).

\begin{remark}
For additional regularization, one may enforce the continuity of physics residuals (e.g., $\nabla \cdot \mathbf{u}$ and momentum residuals) across ghost interfaces.
In this work, we deliberately omit such interface residual constraints to prioritize computational efficiency, as they require evaluating higher-order derivatives on the ghost dataset and substantially increase the autograd graph size and memory footprint.
Consequently, the proposed interface coupling is designed to enforce $C^0$ continuity of the state variables ($\mathbf{u},p$) through $\mathcal{L}_{\text{gh\_u}}$ and $\mathcal{L}_{\text{gh\_p}}$, but does not guarantee higher-order smoothness across sub-domains.
Our empirical studies indicate that enforcing the continuity of $(\mathbf{u},p)$ alone yields sufficiently high reconstruction fidelity, with higher-order interface constraints providing marginal gains in accuracy at a disproportionate computational cost. 
Therefore, unless higher-order regularity is explicitly required by the specific application, we recommend this $C^0$ coupling for optimal scalability. 
The framework remains extensible, allowing users to readily incorporate $C^1$ constraints or PDE residual matching across interfaces if necessary.
\end{remark}

\begin{algorithm}[htbp]
	\caption{Distributed PINNs for Flow Reconstruction}
	\label{alg:dpinn}
	\begin{algorithmic}[1]
		\REQUIRE Global domain $\Omega \times T$; 
        Anchor point $\mathbf{x}_{\text{anc}}$; 
        Weights $\lambda_{\text{obs}}, \lambda_{\text{PDE}}, \lambda_{\text{gh\_u}}, \lambda_{\text{gh\_p}}^{\text{time}}, \lambda_{\text{gh\_p}}^{\text{space\_base}}$;
        Communication interval $I_{\text{comm}}$;
        Number of iterations $N_{\text{iter}}$;
        Parameters for domain decomposition $K, M$.
		\ENSURE Trained local experts $\{\boldsymbol{\theta}_{k,m}^*\}$.
		
		\STATE \textbf{Domain Decomposition:} Partition $\Omega \times T$ into $K \times M$ sub-domains.
		\STATE \textbf{Identify Master Ranks:} $\mathcal{M} \leftarrow \{ (k,m) \mid \mathbf{x}_{\text{anc}} \in \Omega_k \}$.
		\FOR{each sub-domain $(k,m)$}
			\STATE Initialize $\boldsymbol{\theta}_{k,m}$ and local datasets.
			\IF{$(k,m) \in \mathcal{M}$}
				\STATE Set $\lambda_{\text{gh\_p}}^{\text{space}} \leftarrow 0$ \COMMENT{\textit{Master: unidirectional constraint}}
			\ELSE
				\STATE Set $\lambda_{\text{gh\_p}}^{\text{space}} \leftarrow \lambda_{\text{gh\_p}}^{\text{space\_base}}$ \COMMENT{\textit{Slave: alignment active}}
			\ENDIF
		\ENDFOR

		\FOR{iteration $i = 1$ to $N_{\text{iter}}$}
			\IF{$i \pmod{I_{\text{comm}}} \equiv 0$}
				\FOR{each interface with neighbor $j$}
					\STATE Receive $\{\mathbf{u}_{\text{nbr}}\}$.
					\IF{neighbor $j \in \mathcal{M}$}
						\STATE Receive $\{\tilde{p}_{\text{nbr}}\} \leftarrow \{p_j(\mathbf{x},t) - p_j(\mathbf{x}_{\text{anc}}, t)\}$ \COMMENT{\textit{Anchor Normalization}}
					\ELSE
						\STATE Receive $\{p_{\text{nbr}}\} \leftarrow \{p_j(\mathbf{x},t)\}$
					\ENDIF
				\ENDFOR
			\ENDIF
			
			\FOR{each sub-domain $(k,m)$ \textbf{in parallel}}
				\STATE $\mathcal{L} \leftarrow \lambda_{\text{obs}}\mathcal{L}_{\text{obs}} + \lambda_{\text{PDE}}\mathcal{L}_{\text{PDE}} + \lambda_{\text{gh\_u}}\mathcal{L}_{\text{gh\_u}} + \lambda_{\text{gh\_p}}^{\text{space}}\mathcal{L}_{\text{gh\_p}}^{\text{space}} + \lambda_{\text{gh\_p}}^{\text{time}}\mathcal{L}_{\text{gh\_p}}^{\text{time}}$
				\STATE Update $\boldsymbol{\theta}_{k,m} \leftarrow \text{Optimizer}(\mathcal{L})$
			\ENDFOR
		\ENDFOR
		
		\RETURN $\{\boldsymbol{\theta}_{k,m}^*\}$
	\end{algorithmic}
\end{algorithm}

\subsection{Distributed training and hyperparameter adaptation}

To maximize the efficiency and accuracy of our distributed framework, we introduce specific optimizations regarding hardware utilization and heuristic weighting strategy.

\textbf{Hardware acceleration via CUDA Graphs.}
In standard PyTorch implementations, the computational graph for automatic differentiation (autograd) is constructed dynamically at every iteration. 
For PINNs, which necessitate calculating high-order derivatives for physics residuals, this dynamic graph generation incurs significant CPU scheduling overhead. 
This often results in a CPU-bound bottleneck where the GPU remains underutilized, waiting for kernel launches.
To address this, we leverage CUDA Graphs combined with Just-In-Time (JIT) compilation \cite{bafghi2023pinnstorch}. 
Since the topology of the PDE residuals and their derivative chains remains constant during training, CUDA Graphs allow us to capture the complex autograd workflow and compile it into a static computational graph. 
This effectively eliminates the repetitive overhead of rebuilding the derivative graph at each step, bypassing the Python interpreter bottleneck and maximizing computational throughput.

\textbf{Heuristic weighting strategy for loss balancing.}
Balancing the magnitude of different loss terms is crucial for the convergence of multi-objective optimization.
In this work, rather than employing computationally expensive dynamic weighting schemes (e.g., GradNorm), we adopt a physics-motivated heuristic strategy.
We prioritize the reconstruction accuracy within the sub-domains by assigning dominant weights to the observation and PDE terms.
Empirically, we find that stable and accurate reconstruction is generally achieved when the observation term is given higher priority than the physics residual, i.e., $\lambda_{\text{obs}}$ is usually chosen larger than $\lambda_{\text{PDE}}$.
This weighting reflects the typical role of PINNs in flow reconstruction: the network primarily assimilates the available PIV/PTV observations, and the governing equations act as a secondary regularizer to enforce physical consistency and infer unobserved quantities.
Conversely, the weights for ghost layer consistency ($\lambda_{\text{gh\_u}}, \lambda_{\text{gh\_p}}$) are set to relatively lower values compared to the interior losses.
This design is based on two observations.
First, for the velocity components, the local experts are already heavily constrained by the supervised observational data $\mathcal{D}_{\text{obs}}$. 
Consequently, the interface continuity is implicitly encouraged by the data, and the ghost loss serves primarily as a weak constraint to smooth out minor discrepancies at boundaries without conflicting with the ground truth measurements.
Second, regarding pressure, the primary role of $\mathcal{L}_{\text{gh\_p}}$ is to unify the arbitrary integration constants (i.e., level alignment) across sub-domains rather than refining the local pressure gradients (which are governed by $\mathcal{L}_{\text{PDE}}$).
A lower $\lambda_{\text{gh\_p}}$ allows the reference anchor information to propagate globally to correct the baseline offsets, while avoiding stiff gradients that could destabilize the learning of local flow features.

A brief hyperparameter sensitivity discussion (e.g., the communication interval, the number of ghost points, and $\lambda_{\text{gh\_p}}$) is provided in Appendix~\ref{sec:hyperparams-impact}.

\section{Numerical Results}
\label{sec:num_results}

We evaluate the strong scaling performance of our distributed framework by measuring the computational throughput under a fixed global problem size.
Specifically, the total number of observation points $N_{\text{obs}}$ and PDE collocation points $N_{\text{PDE}}$ are held constant across all parallel configurations.
As the number of sub-domains (and corresponding computing processes) increases, the interior data are distributed evenly, reducing the workload per rank approximately linearly.
The network architecture for each local expert is kept identical to ensure consistent representational capacity per sub-domain.
We define the scaling metric based on the wall-clock time per epoch, where an epoch consists of a full pass over the assigned local dataset followed by a single optimizer step.
To handle memory constraints, gradient accumulation is employed to ensure that the mathematical definition of an epoch (and the effective batch size per step) remains invariant.
It is noted that while the interior workload scales ideally, the computational cost for interface consistency (ghost points) grows with the total interface area, introducing a predictable communication and computation overhead as the domain is increasingly partitioned.

In our distributed setting, each subdomain is modeled by an independent local expert, and the global field is assembled by stitching the predictions from all subdomains.
Unless otherwise stated, all quantitative accuracy results are evaluated on all available data points of the corresponding reference dataset using the relative $L^2$ error:
\begin{equation}
\label{eq:relative-l2}
\mathcal{E}(\mathbf{q}) = \frac{\lVert \mathbf{q}_{\mathrm{pred}}-\mathbf{q}_{\mathrm{ref}} \rVert_2}{\lVert \mathbf{q}_{\mathrm{ref}} \rVert_2},
\end{equation}
where $\mathbf{q}$ denotes the quantity of interest (e.g., velocity components).
It is important to note that while the network is trained to minimize the absolute mean squared error (MSE), the metric $\mathcal{E}(\cdot)$ is normalized by the reference magnitude. 
Consequently, variables with smaller norms (e.g., the transverse velocity component $v$ or pressure $p$) may exhibit higher relative errors despite having small absolute deviations.
For the pressure field $p$, which is determined only up to an additive constant, consistent alignment is a prerequisite for accurate evaluation. 
We strictly resolve the gauge freedom based on the role of the subdomain.
For master ranks (subdomains containing the anchor point $\mathbf{x}_{\text{anc}}$), the raw network output retains an arbitrary additive constant. 
We explicitly pin these predictions during post-processing by subtracting the value at the anchor:
\begin{equation}
\tilde{p}_{\mathrm{pred}}(\mathbf{x},t) =
\begin{cases}
    p_{\mathrm{pred}}(\mathbf{x},t) - p_{\mathrm{pred}}(\mathbf{x}_{\text{anc}},t) & (\mathbf{x},t) \in \text{Master Subdomains/Ranks};\\
    p_{\mathrm{pred}}(\mathbf{x},t) & (\mathbf{x},t) \in \text{Slave Subdomains/Ranks}.\\
\end{cases}
\end{equation}
In contrast, for slave ranks, the pressure gauge is implicitly eliminated during the training process via the normalized ghost pressure loss. 
Consequently, their raw network outputs are already aligned to the anchor reference and are used directly without further subtraction.
Subsequently, for both visualization and quantitative error evaluation, we remove the spatial mean from both the stitched prediction and the reference fields to eliminate any global offset:
\begin{equation}
\label{eq:pressure-mean-post}
p'(\mathbf{x},t) \;=\; p(\mathbf{x},t) - \langle p(\cdot,t)\rangle_{\Omega}, \qquad \langle p(\cdot,t)\rangle_{\Omega} := \frac{1}{|\Omega|}\int_{\Omega} p(\mathbf{x},t)\,\mathrm{d}\mathbf{x},
\end{equation}
where $p$ denotes either the anchored prediction $\tilde{p}_{\mathrm{pred}}$ or the reference $p_{\mathrm{ref}}$. 
To ensure statistical robustness, each experimental configuration is executed over five independent trials using a fixed set of random seeds $\{0, 1, 2, 3, 4\}$. 
All reported relative accuracy results are aggregated across these runs and presented in the format of $\text{mean} \pm \text{std}$, representing the arithmetic mean and the standard deviation, respectively.

All experiments are conducted on a single node equipped with up to 8 NVIDIA RTX 5090 GPUs, 720 GB system RAM, and dual Intel(R) Xeon(R) Gold 6459C CPUs with a total of 64 physical cores (128 logical cores).

\subsection{2D steady cavity flow}

We solve the two-dimensional, steady, incompressible Navier-Stokes equations for the classical lid-driven cavity problem on the unit square domain $\Omega=[0,1]^2$ at $\mathrm{Re}=100$. 
The reference solution is available on a $257\times257$ uniform mesh, from which we sample a sparse subset of velocity observations for training. 
Specifically, the observation locations are uniformly sampled on a $10\times10$ grid (i.e., $N_{\text{obs}}=100$).
The experimental setup is summarized in Appendix \ref{sec:hyperparameters}, including the network architecture, the numbers of observation points ($N_{\text{obs}}$) and collocation points ($N_{\text{PDE}}$), and the optimizer configuration.

\begin{table}[t]
    \centering
    \caption{
    Evaluation of the flow reconstruction performance for the 2D steady cavity flow using $P \in \{1, 2, 4\}$ processes: 
    The domain is decomposed such that $P=2$ utilizes a $2 \times 1$ spatial split and $P=4$ utilizes a $2 \times 2$ grid. 
    For each configuration, we report the relative $L^2$ error for the reconstructed velocity and pressure fields.
    To ensure a fair comparison of the distributed architecture, we maintain a fixed number of ghost points per interface while keeping the global number of observation and collocation points constant (thereby reducing the per-rank workload proportionally). 
    }
    \label{tab:cavity2d-scaling}
    \vspace{0.2cm}
    \begin{tabular}{@{}lccc@{}}
        \toprule
        \textbf{Metric} & \textbf{$P=1$} & \textbf{$P=2$} & \textbf{$P=4$} \\
        \midrule
        \multicolumn{4}{c}{\textit{Performance}} \\
        Vel. L2 Error ($\times 10^{-2}$) & $1.46 \pm 0.13$ & $1.33 \pm 0.13$ & $1.29 \pm 0.09$ \\
        Pres. L2 Error ($\times 10^{-1}$) & $1.16 \pm 0.04$ & $1.08 \pm 0.05$ & $1.05 \pm 0.04$\\
        \midrule
        \multicolumn{4}{c}{\textit{Load per Rank}} \\
        PDE Points & $5.0 \times 10^3$ & $2.5 \times 10^3$ & $1.25 \times 10^3$ \\
        Obs. Points & $100$ & $50$ & $25$ \\
        Ghost Points per Interface & --- & $100$ & $100$ \\
        \bottomrule
    \end{tabular}
\end{table}

\begin{figure}[htbp]
    \centering
    \includegraphics[width=0.9\linewidth]{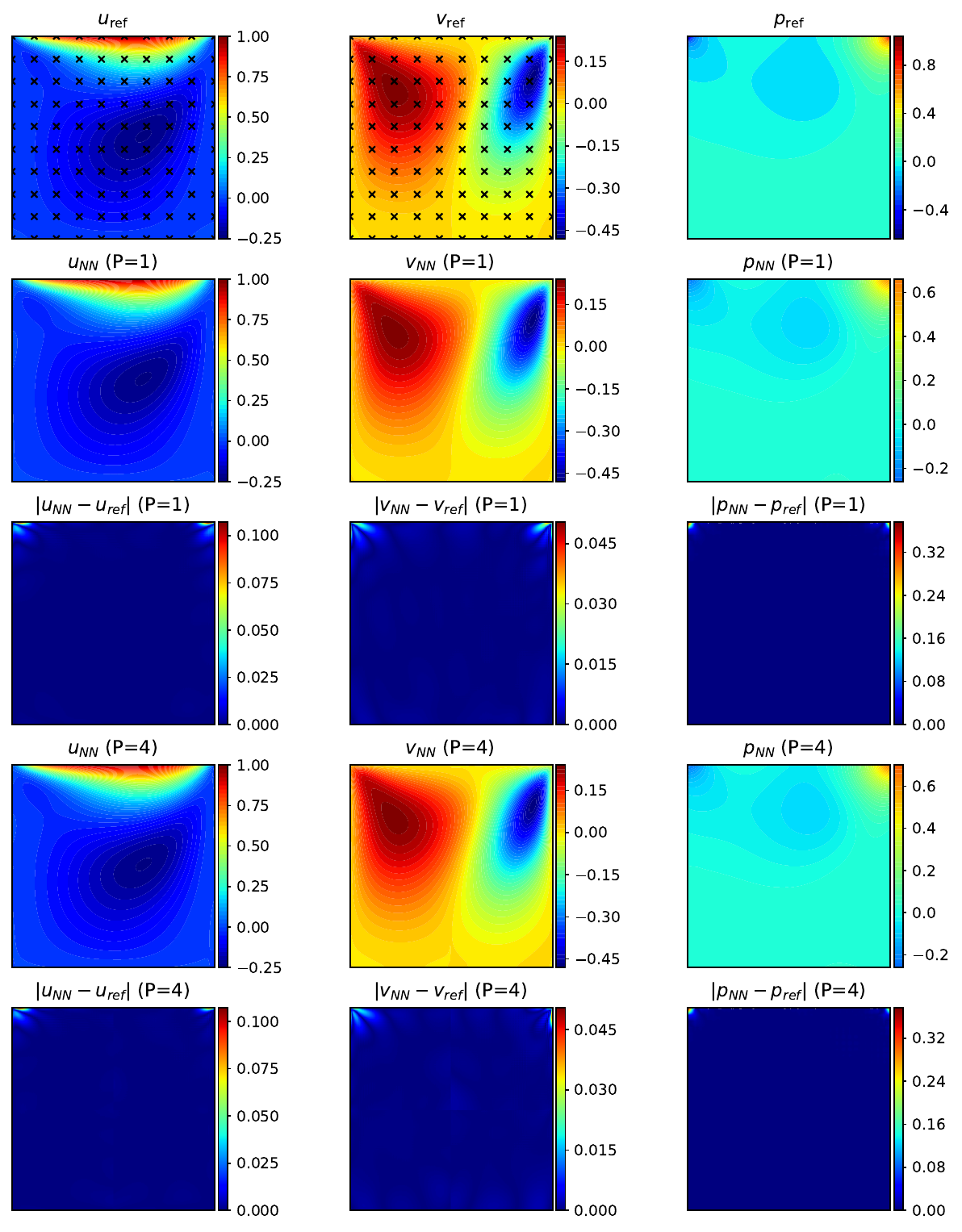}
    \caption{Reconstruction results for the 2D steady lid-driven cavity flow. 
    The panels from top to bottom report the velocity components $(u,v)$ and pressure $p$: reference fields, the $P=1$ (single-domain) PINNs predictions, the corresponding $P=1$ errors, the $P=4$ distributed PINNs predictions with a $2\times2$ spatial decomposition, and the corresponding $P=4$ errors. 
    The ``$\times$'' markers overlaid in the reference $u$ and $v$ panels indicate the locations of the sparse observation points used for training. 
    These results illustrate that both $P=1$ and $P=4$ accurately reconstruct the flow fields.}
    \label{fig:cavity-result}
\end{figure}

To validate the effectiveness of the proposed distributed framework, we compare a single-domain baseline ($P=1$) with distributed configurations $P\in\{2,4\}$, where $P=2$ uses a $2\times1$ spatial decomposition and $P=4$ uses a $2\times2$ spatial decomposition. 
All settings use the same total number of epochs and identical network hyperparameters to ensure a controlled comparison. 
Table \ref{tab:cavity2d-scaling} presents the relative $L^2$ errors alongside the corresponding point distributions. 
As $P$ increases, both the velocity and pressure errors decrease, indicating that under the same global point budget and the same number of training epochs, domain decomposition can improve the reconstruction accuracy.
This accuracy gain suggests that the local experts benefit from focusing on localized flow features within smaller subdomains.
The relatively larger pressure error is expected since $p$ has a smaller characteristic scale and is inferred indirectly from the Navier-Stokes constraints rather than supervised as $\mathbf{u}$, making the relative metric more sensitive.
Figure~\ref{fig:cavity-result} visualizes the reference fields and the reconstructed results for $P=1$ and $P=4$.
The $\times$ markers on the reference $u$ and $v$ fields denote the sparse observation locations.
Since pressure is not directly measured, the reference pressure is provided solely for verification against the inferred predictions.
Both configurations successfully reconstruct the global flow structure, with residual errors primarily confined to the upper cavity corners.
Crucially, the distributed prediction demonstrates seamless stitching, exhibiting no visible discontinuities or artifacts across the sub-domain interfaces.

\begin{figure}[t]
    \centering
    \includegraphics[width=\linewidth]{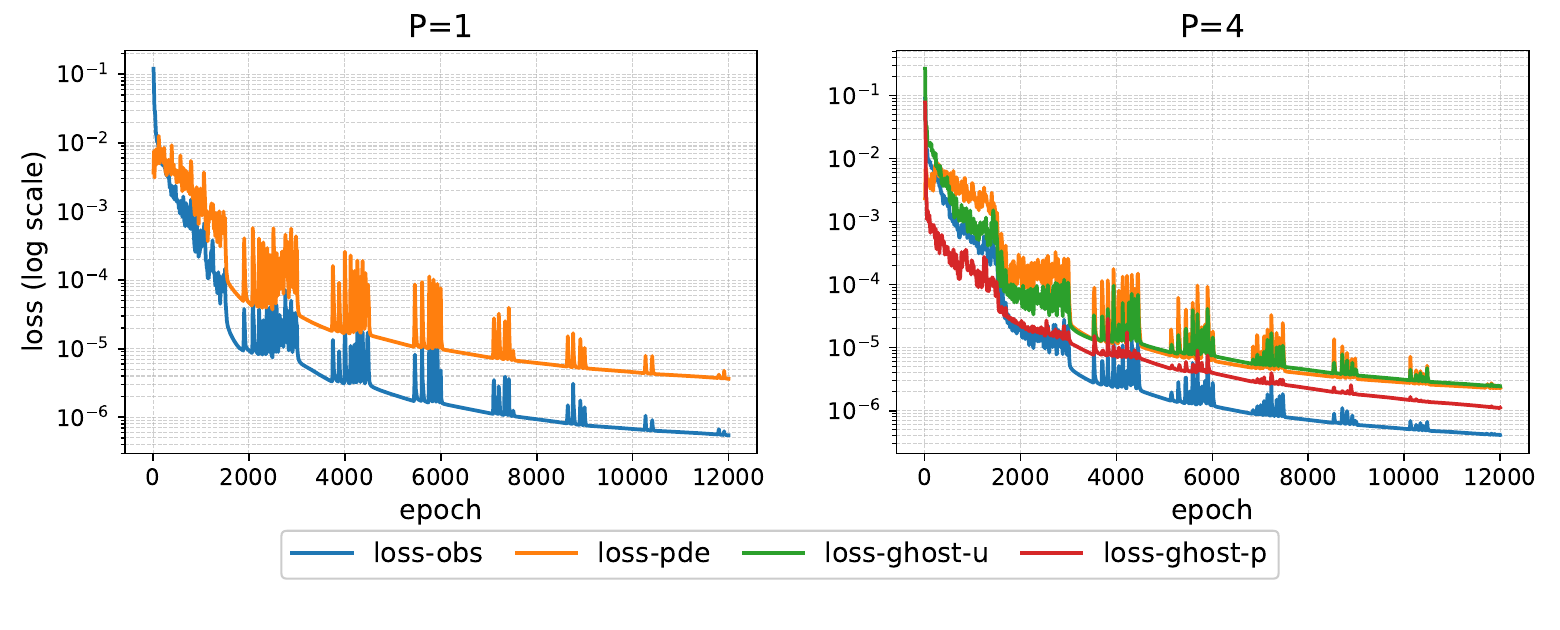}
    \caption{Loss trajectories for the 2D steady cavity flow comparing $P=1$ (single-domain PINN) and $P=4$ ($2\times2$ decomposition). 
    In the $P=1$ case, the two reported curves correspond to the observation loss and the PDE residual loss. 
    In the $P=4$ case, the plotted losses are averaged over the four processes and additionally include the ghost/interface losses for velocity (ghost $u$ loss) and pressure (ghost $p$ loss). 
    The final observation and PDE losses of $P=4$ are lower than those of the $P=1$ baseline, indicating that domain decomposition can effectively improve the overall accuracy.}
    \label{fig:cavity-loss-curve}
\end{figure}

Figure \ref{fig:cavity-loss-curve} further compares the training dynamics. 
Besides the standard observation and PDE losses, the $P=4$ curves (averaged over four ranks) include additional ghost/interface losses introduced to enforce inter-domain consistency. 
Notably, the final observation loss and PDE residual loss for $P=4$ are lower than those of the $P=1$ baseline, suggesting that domain decomposition enables local experts to better fit local flow structures and thereby reduce the overall error.

While this cavity benchmark serves as a straightforward case to validate the accuracy advantages of domain decomposition, it is not ideal for evaluating strong-scaling performance due to its limited global workload.
With a total of only $N_{\text{PDE}}=5.0\times10^3$ collocation points and $N_{\text{obs}}=100$ observations, the computational load per iteration is insufficient to fully utilize the computational throughput of modern GPUs.
Empirically, our experiments showed that GPU utilization rate in this example fluctuated between 10\% and 40\%, confirming that execution efficiency is constrained by computational overhead rather than the raw floating-point arithmetic throughput of the hardware.
To investigate this bottleneck, we conducted detailed profiling on a $P=4$ configuration over a $100$-epoch training window.
The analysis reveals a disparity in execution time: the total CPU time (orchestration and synchronization) exceeds the accumulated GPU kernel time by a factor of approximately $10$.
The workload is highly fragmented with tens of thousands of microsecond-scale kernels, suggesting non-negligible kernel launch and scheduling overheads.
Furthermore, the rank containing the anchor point incurs extra computational work to enforce the pressure gauge (i.e., evaluating the anchor pressure and applying the shift).
As a result, other ranks spend substantial time waiting during ghost-layer data exchanges, which is reflected in the profiler as disproportionately large NCCL synchronization activity on non-anchor ranks.
Crucially, while domain decomposition reduces the number of points per rank, it does not alleviate these fixed scheduling costs.
For instance, as the local expert architecture is kept identical across all ranks, the overhead incurred by parameter updates remains constant regardless of $P$.
In this small-data regime, these invariant costs become the dominant bottleneck, preventing the linear speedup typically achieved in larger-scale benchmark scenarios.

% Fig: 1 process result

\subsection{2D unsteady flow over a cylinder}

We extend our analysis to the unsteady, incompressible Navier-Stokes equations modeling the flow past a circular cylinder at $\mathrm{Re}=100$. 
The computational domain is $\Omega=[-7.50,17.50]\times[-8.00,8.00]$ over the time interval $t\in[0,7.35]$, with the cylinder of radius $R=0.5$ centered at $(0,0)$.
A high-fidelity reference dataset is generated via numerical simulation, providing $(u,v,p)$ across 50 temporal snapshots ($\Delta t=0.15$). 
For training, we use a fixed global budget of $N_{\text{PDE}}=5\times10^5$ collocation points and $N_{\text{obs}}=1\times10^4$ velocity observations, where at each snapshot we randomly select 200 spatial locations within $\Omega$. 
Each sub-domain is modeled by a fully connected MLP with inputs $(t,x,y)$ and outputs $(u,v,p)$, using sinusoidal activations to better represent the periodic wake dynamics. 
To better exploit GPU parallelism and preserve the compute-intensive nature of the training loop, we fix the per-rank batch size to $25{,}000$ for all configurations.
The full hyperparameter configuration is summarized in Appendix \ref{sec:hyperparameters}.

We evaluate strong scaling with $P\in\{1,2,4,8\}$ processes, summarizing the wall time, speedup, and relative $L^2$ error in Table \ref{tab:cylinder2d-scaling}.
Performance profiling reveals that the benchmark is highly compute-bound: the single-process baseline ($P=1$) consumes $\sim 4.3$ GB of GPU memory with sustained utilization in the $81\% \sim 90\%$ range, effectively saturating the hardware.
When scaling to distributed configurations (e.g., $P=4$), the memory footprint increases marginally to $4.5$ GB due to ghost-layer storage, and synchronization overhead causes a modest drop in utilization to $70\% \sim 85\%$.
However, the workload remains sufficiently dense to mask these communication costs.
Consequently, we achieve robust scalability, where each doubling of $P$ consistently yields a speedup exceeding $1.75\times$.
Beyond these computational gains, the distributed approach also enhances solution quality: the domain-wide relative error exhibits a general downward trend as $P$ increases.
This confirms that our distributed PINNs framework achieves a synergy of high-performance computing efficiency and improved reconstruction fidelity.

\begin{table}[t]
    \centering
    \caption{Strong-scaling performance for the 2D unsteady cylinder flow with $P\in\{1,2,4,8\}$ processes. 
    The domain decomposition is configured as: $P=2$ uses a $2\times1$ spatial split, $P=4$ uses a $2\times2$ spatial split, and $P=8$ uses a $2\times2$ spatial split with an additional bisection in time. 
    We report the wall time, the relative $L^2$ error of the reconstructed velocity/pressure field, and the speedup with respect to the previous configuration. 
    In all runs, we keep the number of ghost points per interface fixed, while maintaining the same global totals of observation points and PDE collocation points (thus reducing the per-rank load proportionally). 
    }
    \label{tab:cylinder2d-scaling}
    \vspace{0.2cm}
    \begin{tabular}{@{}lcccc@{}}
        \toprule
        \textbf{Metric} & \textbf{$P=1$} & \textbf{$P=2$} & \textbf{$P=4$} & \textbf{$P=8$} \\
        \midrule
        \multicolumn{5}{c}{\textit{Performance}} \\
        Wall Time & 53m,15s & 30m,15s & 16m,41s & 9m,28s \\
        Speedup (vs. prev) & --- & 1.76 & 1.81 & 1.76 \\
        Vel. L2 Error ($\times 10^{-2}$) & $2.03 \pm 0.10$ & $1.76 \pm 0.12$ & $1.82 \pm 0.09$ & $1.73 \pm 0.04$ \\
        Pres. L2 Error $(\times 10^{-2})$ & $8.43 \pm 0.58$ & $7.79 \pm 0.38$ & $7.89 \pm 0.23$ & $8.03 \pm 0.48$ \\
        \midrule
        \multicolumn{5}{c}{\textit{Load per Rank}} \\
        PDE Points & $5.0 \times 10^5$ & $2.5 \times 10^5$ & $1.25 \times 10^5$ & $6.25 \times 10^4$ \\
        Obs. Points & $1.0 \times 10^4$ & $5 \times 10^3$ & $2.5 \times 10^3$ & $1.25 \times 10^3$ \\
        Ghost Points per Interface & --- & $1.0 \times 10^3$ & $1.0 \times 10^3$ & $1.0 \times 10^3$ \\
        \bottomrule
    \end{tabular}
\end{table}

\begin{figure}[htbp]
    \centering
    \includegraphics[width=0.9\linewidth]{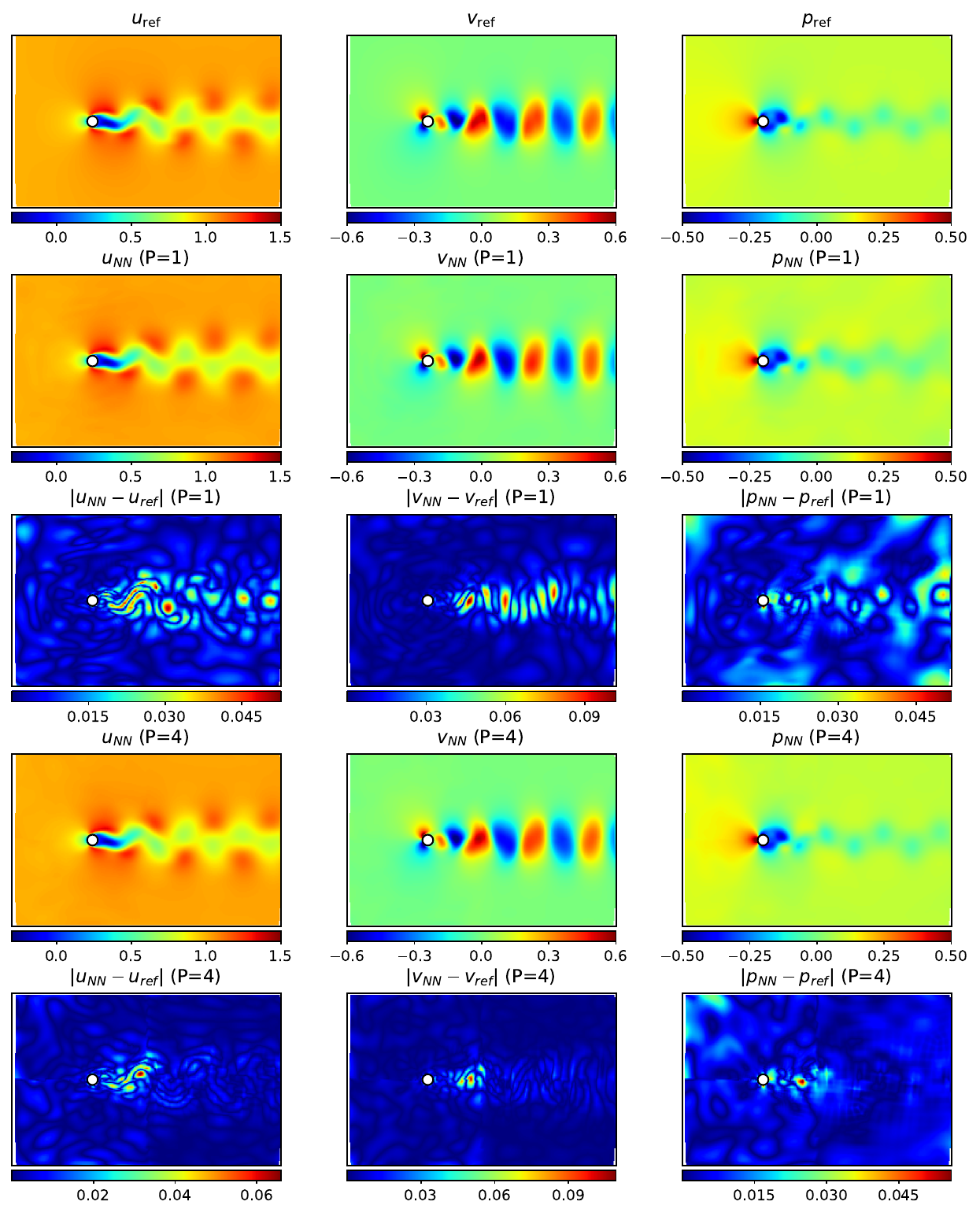}
    \caption{Reconstruction results for the 2D unsteady cylinder wake at $t = 3.75$. The panels compare the reference fields with the $P=1$ (single-domain) PINNs predictions and errors, as well as the $P=4$ distributed PINNs predictions and errors.
    Both $P=1$ and $P=4$ achieve small errors, and the reconstructed $(u,v,p)$ fields remain continuous across the sub-domain interfaces. 
    The results show that the domain decomposition strategy significantly reduces the reconstruction error in the downstream region (the right half). 
    While the $P=1$ struggles to capture the complex wake dynamics uniformly, the distributed approach ($P=4$) achieves a much lower error magnitude in the rear sub-domain.
    }
    \label{fig:cylinder2d-result}
\end{figure}

\begin{figure}[htbp]
    \centering
    \includegraphics[width=\linewidth]{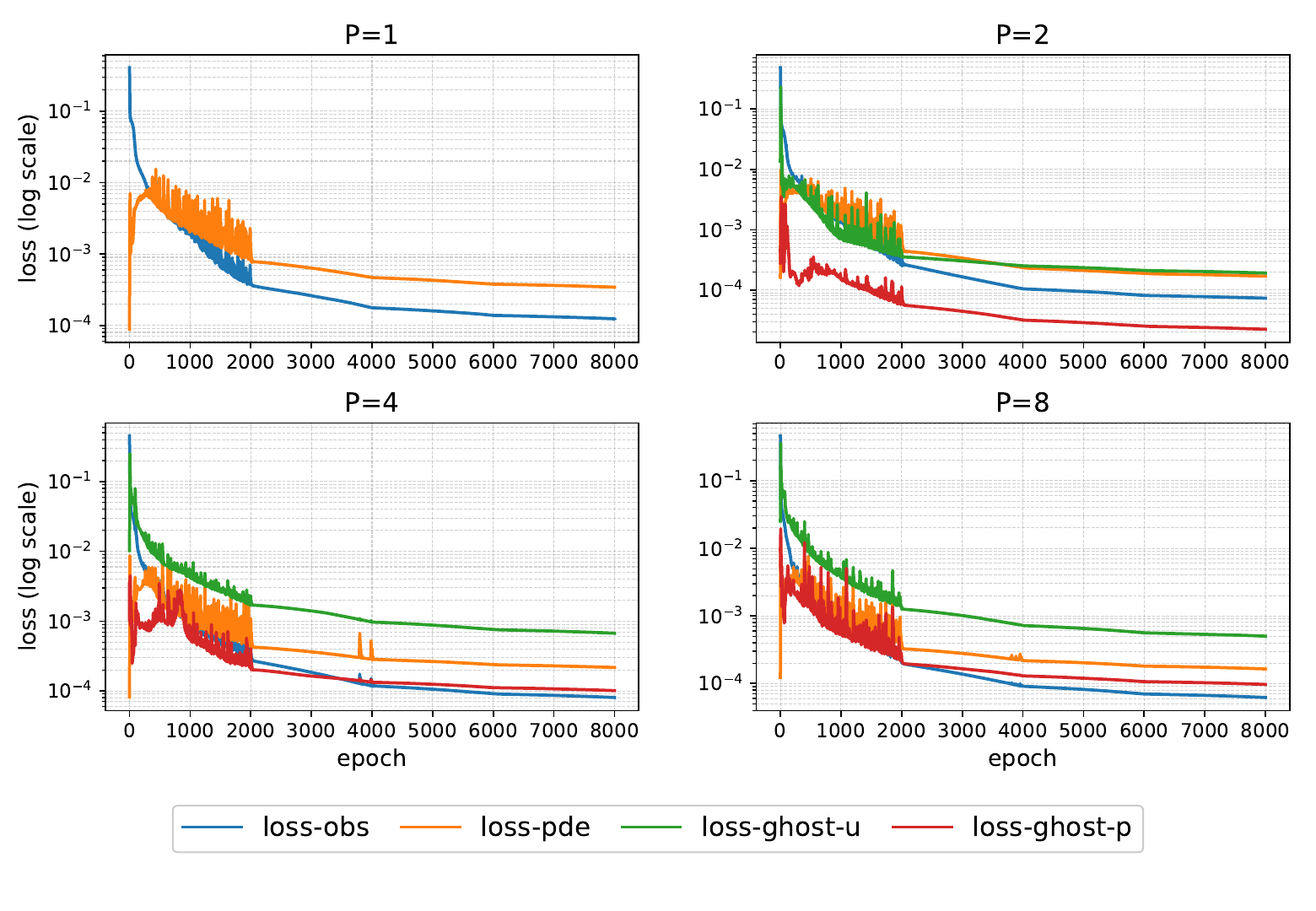}
    \caption{
        Loss trajectories for the 2D unsteady cylinder flow under strong scaling configurations ($P=1,2,4,8$). 
        Across all decompositions, the loss terms decrease monotonically and robustly, with only minor oscillations during the initial training phase. 
        Notably, finer domain decompositions (larger $P$) achieve superior convergence, reaching lower terminal loss values.
        }
    \label{fig:cylinder2d-loss-curve}
\end{figure}

\begin{figure}[htbp]
    \centering
	    \includegraphics[width=\linewidth]{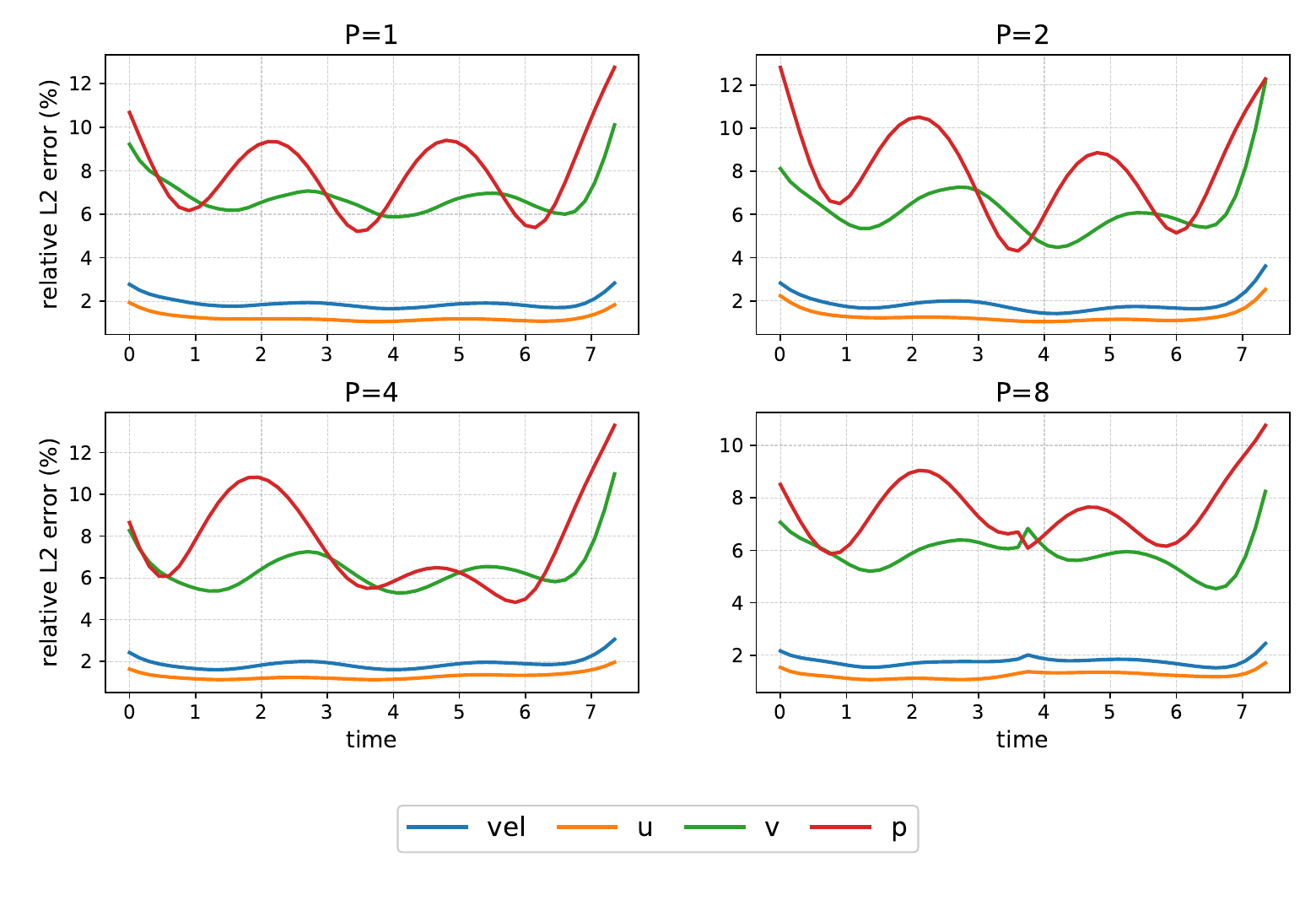}
	    \caption{Relative $L^2$ error evolution over time for the 2D unsteady cylinder flow under strong scaling with $P=1,2,4,8$. 
	    In all configurations, the errors are relatively larger at the beginning and the end of the time window, while remaining well controlled at intermediate times. 
        A slight loss of smoothness may be observed for $P=8$ around the temporal midpoint, which is expected since our interface coupling enforces continuity but not smoothness.
        }
	    \label{fig:cylinder2d-error-over-time}
\end{figure}

Figure \ref{fig:cylinder2d-result} visualizes the reconstruction results at $t = 3.75$, showing reference fields alongside the $P=1$ and $P=4$ predictions and their errors.
While both configurations generally maintain interface continuity, a closer inspection of the error fields reveals distinct differences.
For the single-domain baseline ($P=1$), the error is predominantly localized in the unsteady wake, exhibiting a wavy convective pattern that extends to the outlet.
Notably, the peak errors are concentrated at the vortex cores, exposing the limitation of standard PINNs in resolving high-frequency spatiotemporal dynamics (often referred to as spectral bias).
In contrast, the distributed approach ($P=4$) effectively overcomes this bottleneck.
By decomposing the domain, the sub-networks can focus on local features, thereby capturing the high-frequency information within the vortex cores with much higher fidelity.
Consequently, $P=4$ achieves a significantly lower error magnitude in the rear sub-domain compared to the baseline, validating the effectiveness of domain decomposition in modeling complex, multi-scale flows.
Figure \ref{fig:cylinder2d-loss-curve} further confirms these findings by comparing the training loss histories across $P=1, 2, 4, 8$.
In all cases, the loss trajectories decrease smoothly and stably, indicating robust convergence behavior.
Notably, increasing the number of sub-domains correlates with a lower final loss floor.
Quantitatively, the observation loss for the single-domain baseline ($P=1$) arrives around $1 \times 10^{-4}$, whereas the $P=8$ configuration achieves a significantly lower value of approximately $6 \times 10^{-5}$.
This result demonstrates that domain decomposition facilitates superior fitting capability under the same fixed budget of collocation points and training epochs.
Crucially, it also confirms that the additional inter-domain constraints do not hinder the optimization process; rather, they appear to simplify the learning task for each local sub-network.
Finally, Figure \ref{fig:cylinder2d-error-over-time} reports the time-dependent relative $L^2$ errors.
The error trajectories are stable with only minor fluctuations.
Notably, $u$ and $\mathrm{vel}$ achieve superior accuracy (consistently $<2\%$) compared to $v$ and $p$. 
This is attributed to the larger magnitudes of $u$ and $\mathrm{vel}$, which implicitly weight them higher in the loss minimization process compared to the smaller-scale $v$ and $p$ (which remain bounded within $\sim 12\%$).
We also observe a common trend across all distributed settings where errors marginally increase at the start and end of the time domain.
This phenomenon aligns with the typical behavior of PINNs, reflecting the challenges in strictly resolving temporal boundaries.

\subsection{3D unsteady flow over a cylinder}

We investigate the near wake of a circular cylinder at $\mathrm{Re}=300$ in the fully three-dimensional, unsteady incompressible Navier-Stokes equations. 
The cylinder center is located at $(x,y)=(0,0)$ with radius $R=0.5$, and the cylinder axis is aligned with (and extends along) the $z$-direction. 
The reference dataset consists of high-fidelity CFD snapshots where the flow variables $(u,v,w)$ are discretized on a structured grid. 
We consider a total of 80 snapshots, corresponding to the temporal domain $t\in[0,11.85]$ with a sampling interval $\Delta t=0.15$. 
The spatial domain is $\Omega=[-5,20]\times[-5,5]\times[0,10]$.

Each local partition is modeled by a fully connected MLP with 8 hidden layers of 200 neurons, employing sinusoidal activations to capture the periodic wake dynamics. 
To ensure continuity across sub-domains, interfaces are reinforced with ghost regions of thickness $\delta_{\text{space}}=2.0$ and $\delta_t=2.0$. 
Since the velocity components $(u,v,w)$ exhibit notably different scales in this case, we further apply component-wise weights to the corresponding terms in $\mathcal{L}_{\text{obs}}$ and $\mathcal{L}_{\text{gh\_u}}$ to avoid biasing the network toward any single component; specifically, we use weights $(1,5,100)$.
To improve training stability, we apply gradient clipping with a maximum norm of $1.0$.
Detailed hyperparameter configuration is summarized in Appendix \ref{sec:hyperparameters}.

We evaluate strong scaling with $P\in\{1,2,4,8\}$ processes, summarizing the wall time, speedup, and velocity relative $L^2$ error in Table \ref{tab:cylinder3d-scaling}.
Compared to the 2D benchmark, the 3D benchmark employs larger neural network architectures and a higher volume of training points, rendering the workload highly compute-bound.
Consequently, the computational cost completely dominates communication overheads, resulting in near-ideal scalability.
As evidenced in the table, every doubling of resources yields a speedup factor of approximately $1.9\times$, closely approaching the theoretical limit of $2.0\times$.
Notably, the total training time is drastically reduced from 11h 36m ($P=1$) to merely 1h 40m ($P=8$), representing a nearly $7\times$ acceleration.
Furthermore, the relative $L^2$ error demonstrates a stable and monotonic decrease as domain decomposition increases, validating that our algorithm effectively scales in performance without compromising—and in fact enhancing—accuracy.

\begin{table}[htbp]
    \centering
    \caption{Strong-scaling performance for the 3D unsteady cylinder flow with $P\in\{1,2,4,8\}$ processes. 
    The domain decomposition is configured as: $P=2$ uses a $2\times1\times1$ split, $P=4$ uses a $2\times2\times1$ split, and $P=8$ uses a $2\times2\times2$ split. 
    We report the wall time, the relative $L^2$ error of the reconstructed velocity/pressure field, and the speedup with respect to the previous configuration. 
    In all runs, we keep the number of ghost points per interface fixed, while maintaining the same global totals of observation points and PDE collocation points (thus reducing the per-rank load proportionally). 
    }
    \label{tab:cylinder3d-scaling}
    \vspace{0.2cm}
    \begin{tabular}{@{}lcccc@{}}
        \toprule
        \textbf{Metric} & \textbf{$P=1$} & \textbf{$P=2$} & \textbf{$P=4$} & \textbf{$P=8$} \\
        \midrule
        \multicolumn{5}{c}{\textit{Performance}} \\
        Wall Time & 11h, 36m & 5h,59m & 3h,9m & 1h,40m \\
        Speedup (vs. prev) & --- & 1.94 & 1.90 & 1.89 \\
        Vel. L2 Error $(\times 10^{-2})$ & $2.09 \pm 0.14$ & $1.70 \pm 0.02$ & $1.65 \pm 0.01$ & $1.50 \pm 0.01$ \\
        Pres. L2 Error ($\times 10^{-1}$) & $1.97\pm 0.13$ & $1.58 \pm 0.02$ & $1.48 \pm 0.01$ & $1.39 \pm 0.03$ \\
        \midrule
        \multicolumn{5}{c}{\textit{Load per Rank}} \\
        PDE Points & $6.0 \times 10^5$ & $3 \times 10^5$ & $1.5 \times 10^5$ & $7.5 \times 10^4$ \\
        Obs. Points & $1.0 \times 10^5$ & $5 \times 10^4$ & $2.5 \times 10^4$ & $1.25 \times 10^4$ \\
        Ghost Points per Interface & --- & $5.0 \times 10^3$ & $5.0 \times 10^3$ & $5.0 \times 10^3$ \\
        \bottomrule
    \end{tabular}
\end{table}

\begin{figure}[htbp]
    \centering
    \includegraphics[width=0.9\linewidth]{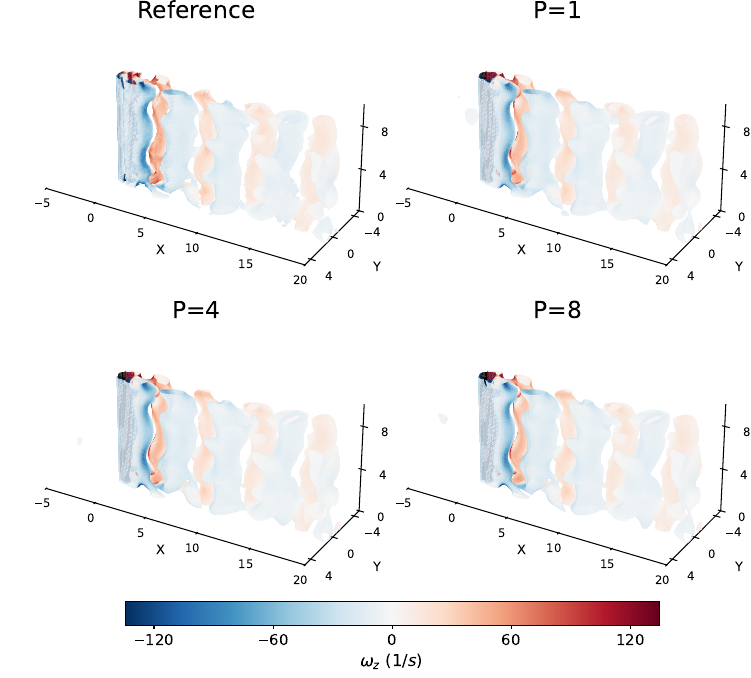}
    \caption{Isosurfaces of the Q-criterion at $t=6$ for the 3D unsteady cylinder wake. 
    The panels compare the reference solution with the reconstructed results obtained using $P=1$, $P=4$, and $P=8$ processes. 
    The proposed distributed PINNs accurately capture the key three-dimensional flow structures, demonstrating faithful recovery of coherent vortical features under increasing domain decomposition.}
    \label{fig:cylinder3d-isosurface}
\end{figure}

\begin{figure}
    \centering
    \includegraphics[width=\linewidth]{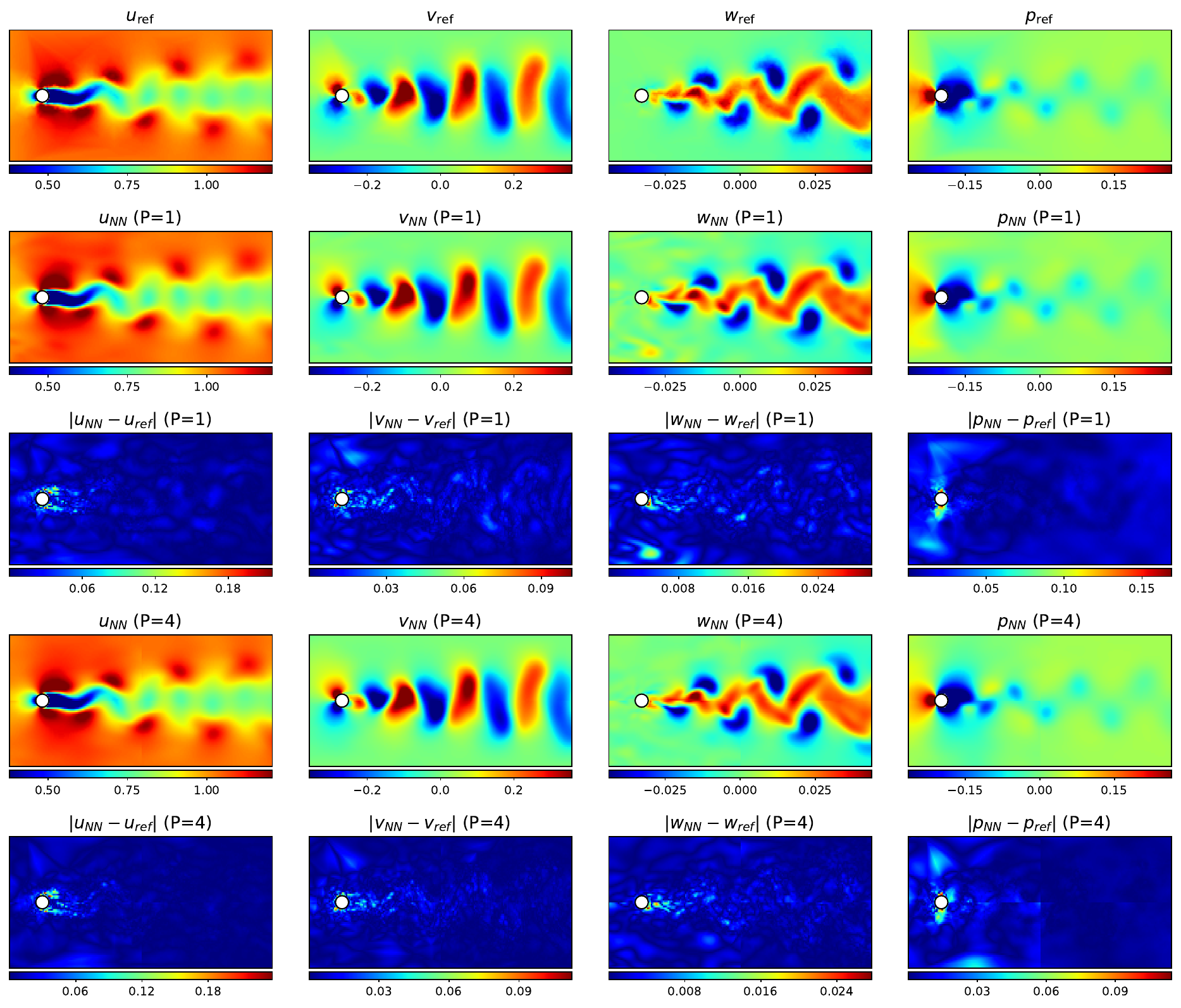}
    \caption{Cross-section plots at $z=0.1$ at $t = 6$ for the 3D cylinder flow. 
    The panels show the reference velocity components $(u,v,w)$ and pressure $p$, the $P=1$ PINN predictions and corresponding errors, and the $P=4$ predictions and corresponding errors. 
    Both PINNs and distributed PINNs capture the fine-scale flow details and accurately reconstruct the pressure field.
    The results show that distributed PINNs significantly reduce the reconstruction error in the downstream region (the right half).
    }
    \label{fig:cylinder3d-result}
\end{figure}

\begin{figure}[h]
    \centering
    \includegraphics[width=\linewidth]{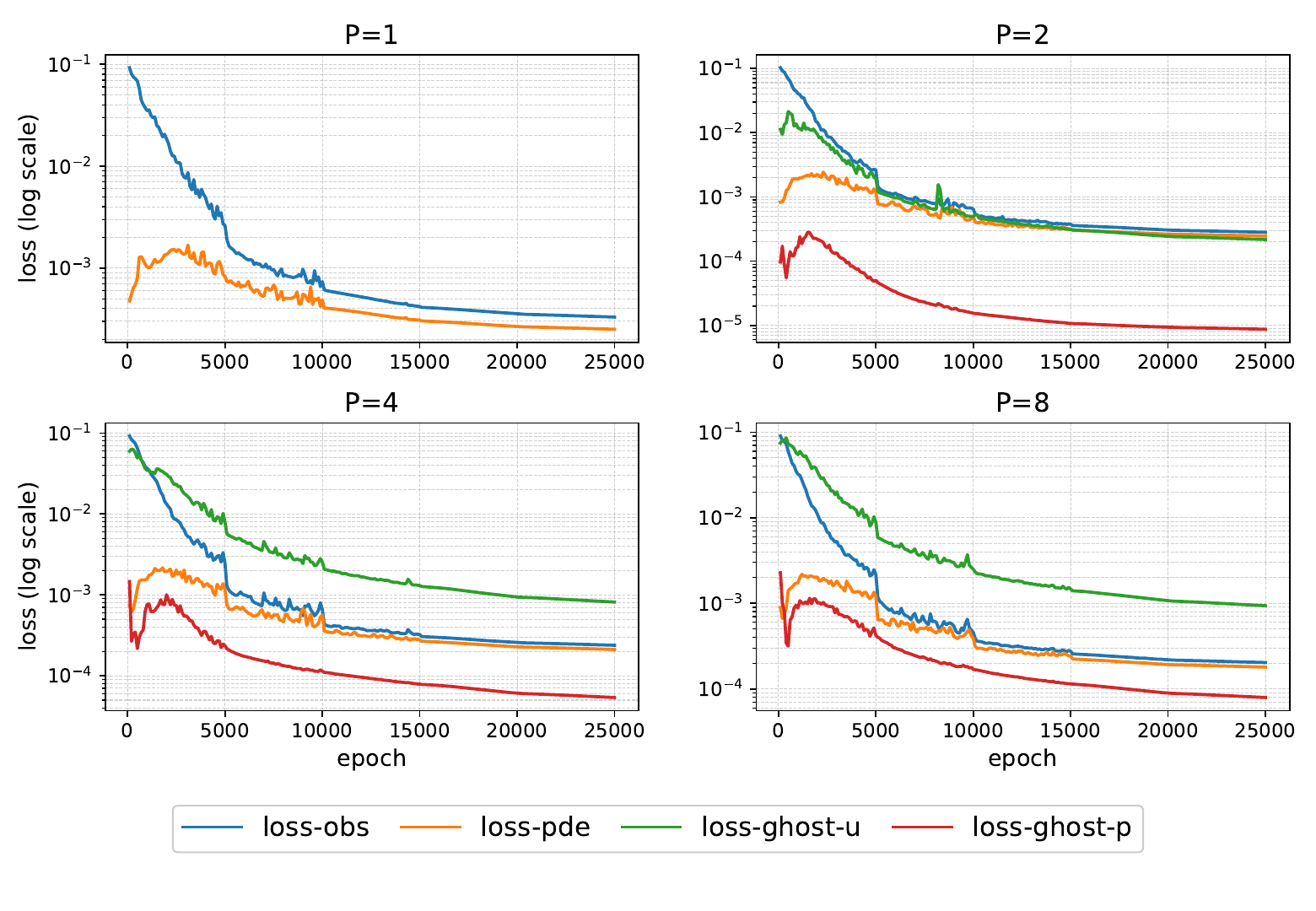}
    \caption{Loss trajectories for the 3D unsteady cylinder flow under strong scaling with $P=1,2,4,8$. 
    The training process converges smoothly and stably across all configurations, and the final loss tends to be smaller as the number of processes increases.}
    \label{fig:cylinder3d-losscurve}
\end{figure}

Figure \ref{fig:cylinder3d-isosurface} visualizes the Q-criterion isosurfaces colored by $z$-vorticity at $t=6$.
This visualization clearly demonstrates that the distributed configurations ($P=4, 8$) faithfully reproduce the complex topology of the wake.
Despite the domain decomposition, the reconstructed flow fields preserve the integrity of the coherent structures and maintain excellent spanwise continuity along the cylinder's axis.
Crucially, the transition across sub-domains appears seamless, with no visible artifacts or discontinuities at the interfaces, confirming that our method effectively handles the spatial dependencies of 3D vortical flows.
A detailed slice-wise comparison is visualized in Figure \ref{fig:cylinder3d-result} ($z=0.1, t=6$).
While both configurations successfully capture the fine-scale features of the flow field, the distributed setting ($P=4$) demonstrates a clear advantage in precision.
As shown in the error sub-plots, the domain decomposition strategy effectively suppresses the reconstruction errors for all variables, yielding a solution that is numerically closer to the ground truth compared to the single-process counterpart.
This validates our method's capability to deliver high-accuracy results even in off-center slices where flow structures are complex.
Figure \ref{fig:cylinder3d-losscurve} depicts the training loss histories across different decompositions.
Consistent with the observations in the 2D benchmarks, the optimization process exhibits stable and smooth convergence behavior for all configurations ($P=1, 2, 4, 8$).
Despite the increased dimensionality and network complexity, the introduction of domain interfaces does not induce numerical instability.
Instead, all loss components—including the governing equation residuals and ghost-layer constraints—decay monotonically, further corroborating the robustness and scalability of our distributed learning framework.

\section{Conclusion}
\label{sec:conclusion}

In this paper, we propose a scalable distributed PINNs framework for flow reconstruction from sparse velocity measurements. 
The global domain is partitioned into multiple sub-domains, each assigned to a local expert, with overlapping ghost layers ensuring physical continuity across sub-domain interfaces.
To ensure global physical consistency in the distributed inverse setting, we introduce reference anchor normalization with decoupled asymmetric weighting, which eliminates pressure gauge indeterminacy and aligns pressure across spatial interfaces and temporal boundaries without introducing additional gauge-fixing loss terms and with only limited overhead. 
We further develop a hardware-efficient training pipeline based on CUDA Graphs and JIT compilation to alleviate the CPU-side overhead of high-order autodifferentiation in PyTorch and improve per-GPU throughput. 
Numerical experiments on a 2D steady lid-driven cavity, a 2D unsteady cylinder wake, and a 3D unsteady cylinder wake validate both accuracy and scalability. 
By monitoring wall time and speedup across all cases, we observe that our method yields high-fidelity reconstructions and demonstrates near-linear strong scaling. 
These speedups become more pronounced as problem size and flow complexity increase, primarily because the computational cost of the forward pass—the primary focus of our distributed acceleration—becomes the dominant bottleneck, effectively diminishing the relative overhead of backpropagation and inter-process communication.

In summary, our primary contributions include the development of a distributed PINNs framework specifically tailored for fast flow reconstruction, the introduction of a novel normalization strategy to resolve pressure indeterminacy in distributed inverse problems, and the integration of hardware-level optimizations to ensure efficient scalability for large-scale, complex flow benchmarks. The framework developed in this work establishes a scalable and engineerable pathway for the reconstruction of flow fields and the understanding of complex hydrodynamic phenomena.

\section*{Data availability}
Data and codes will be made available upon publication.

\section*{Declaration of competing interest}
The authors declare that they have no known competing financial interests or personal relationships that could have appeared to influence the work reported in this paper.






% \end{thebibliography}

% \newpage

\appendix

\setcounter{table}{0}
\setcounter{figure}{0}

\section*{Appendix A. Hyperparameters Impact Analysis}
\label{sec:hyperparams-impact}

\textbf{Impact of Communication Interval.}
The communication interval controls how often neighboring subdomains exchange interface information. 
A larger interval (fewer communications) can improve efficiency, but the benefit depends on whether runtime is dominated by communication or computation. 
When the per-step compute workload is small, frequent communication can become a bottleneck and significantly reduce throughput; 
conversely, for computation-heavy cases, reducing communication has limited impact on wall time.
The interval also affects optimization.
Overly sparse communication can cause subdomains to drift and accumulate larger interface mismatches. 
Empirically, an effective strategy is to select the interval based on 
(i) the dominant bottleneck of the benchmark and 
(ii) the loss-curve stability, especially the presence/absence of persistent oscillations in ghost/interface losses. 
A good default is the largest interval that avoids noticeable oscillations while not being communication-limited.

\textbf{Impact of Number of Ghost Points.}
The number of ghost points should be chosen according to how well each local expert generalizes within the ghost layer. 
A practical diagnostic is to compare (i) the mismatch on the \emph{selected} ghost points used for training and (ii) the mismatch on additional points randomly sampled in the same ghost layer. 
If the selected ghost points show small discrepancies while the randomly sampled points still exhibit large errors, the ghost-point set is likely too sparse and does not adequately cover the interface region. 
In contrast, if the mismatch is already large on the selected ghost points, the issue is typically insufficient enforcement rather than sampling density, suggesting that $\lambda_{\text{gh\_u}}$ and/or $\lambda_{\text{gh\_p}}$ may be too small.

\textbf{Impact of $\lambda_{\text{gh\_p}}$.}
The pressure ghost term is mainly used to align the pressure gauge across subdomains (i.e., remove the arbitrary constant offset of local pressure solutions) and is not the primary driver of pressure continuity. 
In many cases, once $(u,v,w)$ are well matched across interfaces and the Navier-Stokes residual is minimized, the pressure field shape is constrained accordingly and should be consistent up to a constant.
Therefore, when pressure shows localized discontinuities near interfaces (rather than a block-wise shift), it is usually more effective to strengthen the velocity/interface and physics constraints (e.g., tuning $\lambda_{\text{gh\_u}}$, $\lambda_{\text{PDE}}$, and the velocity observation weight) instead of increasing $\lambda_{\text{gh\_p}}$.

\section*{Appendix B. Hyperparameters for Numerical Experiments}
\label{sec:hyperparameters}

Table \ref{tab:hyperparams} summarizes the hyperparameter configurations used in all numerical experiments. For each benchmark, we report the network architecture, training settings, dataset sizes (observation and collocation points), and the domain-decomposition/interface parameters, providing a complete specification for reproducing the reported results.

\renewcommand{\thetable}{B\arabic{table}}
\begin{table}[htbp]
    \centering
    \caption{Hyperparameter configurations for numerical experiments. Columns report the settings used for the 2D steady cavity flow, the 2D unsteady cylinder flow, and the 3D unsteady cylinder flow, respectively.}
    \label{tab:hyperparams}
    \vspace{0.2cm}
    {\small
    \renewcommand{\arraystretch}{1.15}
    \begin{tabular}{@{}l p{0.2\linewidth} p{0.2\linewidth} p{0.2\linewidth}@{}}
        \toprule
        \textbf{Parameter} & \textbf{2D cavity} & \textbf{2D cylinder} & \textbf{3D cylinder} \\
        \midrule
        \multicolumn{4}{c}{\textit{Problem \& Network Architecture}} \\
        Reynolds Number ($Re$) & $100$ & $100$ & $300$ \\
        Hidden Layers & $6$ & $6$ & $8$ \\
        Neurons per Layer & $80$ & $150$ & $200$ \\
        Activation Function & $\tanh(\cdot)$ & $\sin(\cdot)$ & $\sin(\cdot)$ \\
        Outputs & $(u, v, p)$ & $(u, v, p)$ & $(u, v, w, p)$ \\
        \midrule
        \multicolumn{4}{c}{\textit{Training Configuration}} \\
        Max Epochs & $12{,}000$ & $8{,}000$ & $25{,}000$ \\
        Batch Size per Rank & $1{,}250$ & $25{,}000$ & $25{,}000$ \\
        Learning Rate & $1\times10^{-2}$ & $1\times10^{-3}$ & $1\times10^{-3}$ \\
        LR Factor \& Interval & $0.5$, $1.5k$ epochs & $0.2$, $2k$ epochs & $0.3$, $5k$ epochs \\
        \midrule
        \multicolumn{4}{c}{\textit{Dataset \& Sampling}} \\
        Obs Points $N_{\text{obs}}$ & $100$ & $10{,}000$ & $100{,}000$ \\
        PDE Points $N_{\text{PDE}}$ & $5{,}000$ & $500{,}000$ & $600{,}000$ \\
        Ghost Points per Interface & $100$ & $1{,}000$ & $5{,}000$ \\
        \midrule
        \multicolumn{4}{c}{\textit{Distributed Setup \& Loss Weights}} \\
        Interface Thickness & $\delta_{xy}=0.2$ & $\delta_{xy}=2.0,\,\delta_{t}=1.0$ & $\delta_{xyz}=2.0,\,\delta_{t}=2.0$ \\
        Communication Interval & $1$ epoch & $1$ epoch& $1$ epoch \\
        $(\lambda_{\text{obs}},\lambda_{\text{PDE}},\lambda_{\text{ghost},u},\lambda_{\text{ghost},p})$ & $(10.0,\,4.0,\,1.0,\,1.0)$ & $(10.0,\,5.0,\,1.0,\,1.0)$ & $(10.0,\,10.0,\,1.0,\,1.0)$ \\
        \bottomrule
    \end{tabular}
    }
\end{table}

% \end{linenumbers}

\bibliographystyle{elsarticle-num} 
\bibliography{cas-refs.bib}

\end{document}